\documentclass[letterpaper]{article} 
\usepackage{aaai24}  
\usepackage{times}  
\usepackage{helvet}  
\usepackage{courier}  
\usepackage[hyphens]{url}  
\usepackage{graphicx} 
\usepackage{natbib}  
\usepackage{caption} 
\usepackage{algorithmic}

\usepackage{algorithm}

\usepackage{newfloat}
\usepackage{listings}
\DeclareCaptionStyle{ruled}{labelfont=normalfont,labelsep=colon,strut=off} 
\floatstyle{ruled}
\newfloat{listing}{tb}{lst}{}
\floatname{listing}{Listing}
\nocopyright
\usepackage[utf8]{inputenc} 
\usepackage[T1]{fontenc}    
\usepackage{url}            
\usepackage{booktabs}       
\usepackage{amsfonts}       
\usepackage{nicefrac}       
\usepackage{microtype}      
\usepackage{xcolor}    
\usepackage{graphicx}
\usepackage{subcaption}
\usepackage{amsmath}
\usepackage{bm}
\usepackage{algorithm}
\usepackage{algorithmic}
\usepackage{multirow}
\usepackage{enumitem}
\usepackage{caption}
\title{Unbiased Decisions Reduce Regret: Adversarial Domain Adaptation for the Bank Loan Problem}
\author {
    Elena Gal\textsuperscript{\rm 1},
    Shaun Singh\textsuperscript{\rm 2},
    Aldo Pacchiano\textsuperscript{\rm 3},
    Ben Walker \textsuperscript{\rm 4},
    Terry Lyons \textsuperscript{\rm 4}
    Jakob Foerster \textsuperscript{\rm 5}
}
\affiliations {
    \textsuperscript{\rm 1}Institute of Biomedical Engineering, Dept. Engineering Science, University of Oxford\\
    elena.gal@eng.ox.ac.uk\\
    \textsuperscript{\rm 2}Conduit\\sdsingh@conduit.dev\\
    \textsuperscript{\rm 3}Broad Institute of Harvard and MIT\\pacchiano@berkeley.edu\\
    \textsuperscript{\rm 4} Mathematical Institute, University of Oxford\\
    walkerb1@maths.ox.ac.uk\\
    tlyons@maths.ox.ac.uk\\
     \textsuperscript{\rm 5} FLAIR, University of Oxford\\
    jakobfoerster@gmail.com\\
}

\usepackage{verbatim}
\usepackage{amsmath}
\usepackage{amssymb}
\usepackage{amsthm}
\usepackage{comment}

\usepackage{mathtools}
\usepackage{bm}
\usepackage[utf8]{inputenc}
\usepackage{tikz-cd}
\usepackage{wasysym}
\usepackage{varwidth}
\usepackage{mathtools}
\usepackage{xspace}
\usepackage{aliascnt}
\usepackage{manfnt}
\usepackage{mathrsfs}
\usepackage[bbgreekl]{mathbbol}
\usepackage[title,toc]{appendix}

\newcommand{\R}{\mathbb{R}}

\newcommand{\bae}{\begin{equation}\begin{aligned}}
\newcommand{\eae}{\end{aligned}\end{equation}}
\newcommand{\beq}{\begin{equation}}
\newcommand{\eeq}{\end{equation}}

\begin{document}

\maketitle

\begin{abstract}
In many real world settings binary classification decisions are made based on limited data in near real-time, e.g. when assessing a loan application. We focus on a class of these problems that share a common feature: the true label is only observed when a data point is assigned a positive label by the principal, e.g. we only find out whether an applicant defaults if we \emph{accepted} their loan application. As a consequence, the \emph{false rejections} become self-reinforcing and cause the labelled training set, that is being continuously updated by the model decisions, to accumulate bias. Prior work mitigates this effect by injecting optimism into the model, however this comes at the cost of increased false acceptance rate. We introduce adversarial optimism (AdOpt) to directly address bias in the training set using adversarial domain adaptation. The goal of AdOpt is to learn an unbiased but informative representation of past data, by reducing the distributional shift between the set of accepted data points and all data points seen thus far. AdOpt significantly exceeds state-of-the-art performance on a set of challenging benchmark problems. Our experiments also provide initial evidence that the introduction of adversarial domain adaptation improves fairness in this setting.
\end{abstract}
\label{sec:intro}

In a variety of online decision making problems, principals have to make an acceptance or rejection decision for a given instance based on observing data points in an online fashion. Across a broad range of these, the \textit{true label} is only revealed for those data points which the principal accepted, creating a biased labelled dataset. 

In this work we concentrate on addressing this issue for the specific class of binary classification tasks, also known as the ``Bank Loan Problem''(BLP), motivated by the characteristic example of a lender deciding on outcomes of loan applications. The lender's objective is to maximize profit, i.e. accept as many credible applicants as possible, while denying those who would ultimately default. The caveat is that the lender doesn't learn whether rejected applicants would have actually repaid the loan. Hence a decision policy that relies solely on the past experience lacks the opportunity to correct for erroneous rejection decisions. These ``false rejects'' are self-reinforcing since the correct label is never revealed for rejected candidates.

As time progresses, the growing pool of accepted applicants (the models training set), created by the models decisions, forms an increasingly biased dataset, whose distribution is different from that of the general applicant population. This distributional shift affects the accuracy of the predictions of a model trained on the set of accepted points for any future applicants. 

We propose and evaluate a novel approach to the BLP that is motivated by methods from the domain of learning in the presence of distributional shift. Specifically, we utilize \emph{adversarial domain adaptation} to learn a de-biased representation of the training data that minimizes this distributional difference, while preserving the informative features. Although very natural in this setting, this is to our knowledge the first attempt to utilise adversarial domain adaptation to tackle bias in the online context. 

Our experiments show that the \emph{adversarially de-biased} classifier trained on the above ``domain-agnostic'' representation of the data achieves increased recall on the new queries, and thus mitigates the issue of self-reinforcing false rejections.
However, on its own the adversarially de-biased classifier suffers from a fundamental flaw: It needs to trade-off between \textit{truly informative} features and reducing bias. Clearly, both objectives cannot be perfectly accomplished at the same time, which causes this method to perform poorly on some datasets. To overcome this issue we propose adversarial optimism method (AdOpt), which combines the de-biased classification approach with the recently proposed Pseudo-Label Optimism(PLOT)~\cite{PLOT}. PLOT is a strategy for injecting optimism into a model by adding each incoming query to the dataset with a positive pseudo-label and re-training an \textit{optimistic classifier} on the resulting dataset. However when this strategy is applied indiscriminately to all incoming queries it suffers from greatly increased \textit{false acceptance} rate. AdOpt proposes an alternative to this approach by first filtering the incoming queries using the adversarially de-biased classifier to assess the probability of a query being a true positive. In other words, AdOpt proposes a data driven strategy for selecting pseudo-label candidates. 

\begin{figure*}
  \centering
  {\includegraphics[width=1\linewidth]
  {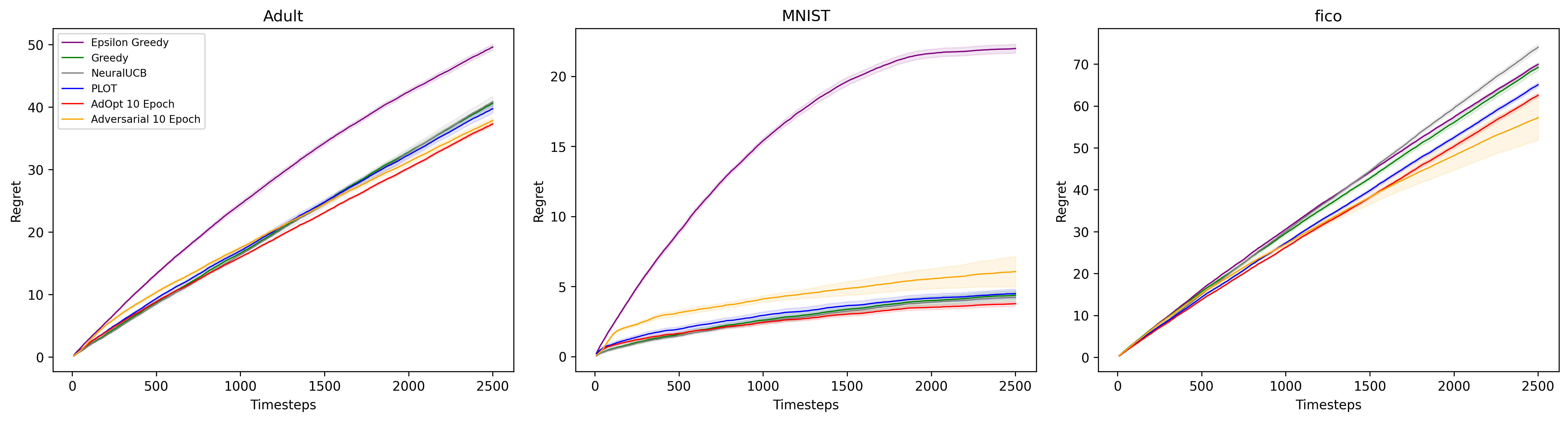}}
  \caption{Cumulative regret of AdOpt  vs. PLOT, 
  NeuralUCB, Greedy and Epsilon-greedy methods on the UCB dataset \emph{Census Income (``Adult'')} as well as MNIST and FICO challenge dataset(``FICO''). We also evaluate the adversarially de-biased classifier on its own as an ablation (``Adversarial''). The adversarial component is trained for 10 epoch at each batch. Cumulative regret is reported as a function of timestep (batch) and expressed in multiples of batch size, i.e. one unit of regret on the graph signifies 32 misclassifications. The shaded region represents one standard deviation from the mean, computed across 25 experiments, 5 runs over 5 different sampling methods. We find that AdOpt outperforms the previous SOTA methods by a significant margin while displaying stable performance for all datasets and seeds evidenced by its narrow STD interval.
  }
  \label{fig:regrets}
\end{figure*}

Since the de-biased classifier is able to choose positives with better accuracy, its utilisation allows AdOpt to converge after seeing a smaller number of candidates and with fewer misclassified queries, as evidenced by its lower \emph{regret} in our experiments. AdOpt consistently outperforms other established approaches from the literature, namely PLOT, ``greedy'', $\epsilon$-greedy (with a decaying schedule) and NeuralUCB~\cite{zhou2020neural} algorithms (see Figure ~\ref{fig:regrets}). Conducting high number of experiments across several sampling methods allows us to confirm the statistical significance of our results with high accuracy (Figures ~\ref{fig:regrets}, ~\ref{fig:PLOTAdOpt}). 

The term ``de-biased classifier'' in our context means a classifier that is trained on a representation of the data that diminishes differences between the accepted applicants and general population, hence a priori de-biasing in our sense doesn't necessarily imply increased  ``fairness'' e.g. towards applicants with protected characteristics. However we conduct some initial experiments to clarify whether that is the case. As Figure~\ref{fig:fairness} illustrates, the adversarially de-biased classifier when used on its own in the BLP setup comes closer than the established approaches to satisfying the \emph{equality of odds} definition of fairness. As a result AdOpt that incorporates adversarial de-biasing also performs better on this measure compared to PLOT. Remarkably, this increase in fairness is not due to explicitly controlling for bias towards protected characteristics, which usually comes at a trade-off with classification accuracy. To our knowledge our work is the first to evaluate fairness in the context of BLP. Our results indicate that incorporating adversarial de-biasing has potential to increase fairness, and proposes a potentially fruitful and socially relevant new direction for investigation.

\section{Related work}

\paragraph{Contextual bandits and function approximation}
The Bank Loan problem (BLP) is a specific instance of a contextual bandit problem with two possible actions and a reward function that is 1 if the accepted applicant returns the loan, -1 if he doesn't and 0 if the lender decides to reject the applicant. There exist a multitude of methods for this setting ranging from adversarial bandit approaches such as EXP4~\cite{auer2002nonstochastic} to methods that rely on simple parametric assumptions on the data (for example linear responses) such as OFUL~\cite{abbasi2011improved} or GLM~\cite{glm_bandit}. These methods' applicability to the BLP depends on the data satisfying restrictive parametric assumptions. Specifically, these methods are not compatible with the use of deep neural network (DNN) function approximators. 

\paragraph{Neural solutions to the bank loan problem}
More recently, deep neural networks have been successfully applied to this class of problems, with several proposed strategies~\cite{riquelme2018deep, zhou2020neural,xu2020neural}. For example, in~\cite{riquelme2018deep} the authors conduct an extensive empirical evaluation of existing bandit algorithms on public datasets. A more recent line of work has focused on devising ways to add optimism to the predictions of neural network models. Methods such as Neural UCB and Shallow Neural UCB~\cite{zhou2020neural,xu2020neural} are designed to add an optimistic bonus to the model predictions that is a function of the representation layers of the model. Their theoretical analysis is inspired by insights gained from Neural Tangent Kernel theory. Other recent works in the contextual bandit literature such as~\cite{foster2020beyond} have started to pose the question of how to extend theoretically valid guarantees into the function approximation scenario, so far with limited success~\cite{sen2021top}. 

The PLOT method from~\cite{PLOT} provides a simple and computationally efficient way to introduce optimism which can be used in combination with deep neural networks. This optimism translates \textit{self-fulfilling} false rejects into \textit{self-correcting} false accepts by the way of introducing pseudo-labelled queries to the model training set. However, this approach suffers from increased false acceptance rate in particular early on in the learning process. Therefore while the above approach has the benefit of theoretical performance guarantees, to obtain SOTA empirical results in~\cite{PLOT} the authors mitigate this problem by assigning pseudo-labels with fixed (small) probability $\epsilon$. However such one-size-fits-all approach doesn't take into account the intrinsic properties of the dataset or a specific query which hinders performance when comprehensively evaluated across several sampling methods. In the present work evaluation across several seeds on the MNIST and Credit datasets shows that on average PLOT ties with NeuralUCB, whereas AdOpt outperforms both by a significant margin, as evidenced by the relative t-values we report in Table~\ref{tbl:tvalues}. 
\paragraph{Selective labels} The bank loan problem is also known as the``selective labels'' problem~\cite{lakkaraju2017selective}, but the majority of the literature in this area so far focuses on the offline setup. The recent paper~\cite{wei2021decision} considers online scenario, but in contrast with our setting their online mechanism requires the data distribution to be stationary. Additionally although the authors mention their methods can be extended to the setting of NNs, their experimental evaluation is conducted mainly over linear models. 
\paragraph{Reject inference} This line of methods attempts to correct bias in credit scoring by modelling the missing outcomes of the rejected applications. While statistical and machine learning methods are most commonly used, more recently deep learning methods have also been applied~\cite{mancisidor2020deep}. However, like in the majority of selective labels literature, the online scenario has not been considered.
\paragraph{Fairness} \label{lit:fairness} Fairness is a central concern in algorithmic decision-making and existing literature offers numerous approaches and perspectives. Fairness in the contextual bandits setting is explored in~\cite{joseph2016fairness, schumann2019group}. Domain adaptation techniques for censoring of protected features were proposed and evaluated by~\cite{Edwards2015Censoring, zhang2018mitigating} among others and~\cite{coston2019fair, singh2021fairness} address fairness in the context of distributional shift. The above works do not consider the online setting, or more specifically the BLP. In the setup of ``selective labels''~\cite{kilbertus2020fair} considers learning under fairness constraints under assumption of stationary data distribution (not required by our method) and~\cite{coston2021characterizing} proposes a method for selecting fair models from existing ones. 
\paragraph{Repeated Loss Minimization} Previous works~\cite{hardt2016strategic, hashimoto2018fairness} have studied the problem of repeated classification settings where the underlying distribution is influenced by the model itself. The works~\cite{perdomo2020performative, miller2021outside}  extend this idea to introduce the problem domains of \emph{strategic classification} and \emph{performative prediction}. Unlike in the above setting, in our case only the labelled training set is affected by the model, leading to a distributional shift \emph{between} this set and the set of points presented to the model. This shift is precisely the reason for suitability of \emph{domain adaptation} to our setting.
\section{Background}
\label{sec:background}
\subsection{The bank loan problem}
\label{sec:BLP}
We are focusing on the class of sequential learning tasks, where at every timestep $t$, the learner is presented with a query $\mathbf{x}_t\in \mathbb{R}^d$ and needs to predict its binary label $y_t\in \{ 0,1\}$. Acceptance of the query carries an unknown reward of $2y_t -1$ and rejection a fixed reward of $0$. Using the bank loan analogy, the lender receives a unit gain for accepting a customer who repays his loan, experiences a unit loss for accepting a customer who defaults and receives nothing if they reject an application. The learners objective is to maximize their reward, which is equivalent to maximizing the number of correct decisions made during training ~\cite{kilbertus2020fair}. 

We assume that the distribution of the labels is given by a probability function $\rho_{\theta_{\star}}$:

\begin{equation}\label{equation::glm_assumption}
    y(\mathbf{x_t}) = \begin{cases}
            1 & \text{with probability } \rho_{\theta_\star}(\mathbf{x}_t)\\
            0 & \text{o.w.}
        \end{cases}
\end{equation}
where $\rho_{\boldsymbol{\theta}_\star}: \mathbb{R}^d \rightarrow \mathbb{R}$ is a function parameterized by $\boldsymbol{\theta}_\star \in \Theta$. Note that the distribution of the \emph{query points} $\mathcal{P}(\mathbf{x})$ may be different for different $t$'s: e.g. the bank may extend it's loan provision at a certain time point $t_0$ and start targeting a different applicant demographic. 

We assume that the learners decision at time $t$ is parameterized by $\boldsymbol{\theta}_t$ and is given by the rule:
\begin{equation*}
 \text{If }   \rho_{\boldsymbol{\theta}_t}(\boldsymbol{x}_t) \geq \frac{1}{2} \text{ accept } 
\end{equation*}
A common measure of the learners performance is \emph{regret}:
\begin{equation*}\label{eq:regret}
    \mathcal{R}(t) = \sum_{\ell=1}^t \max(0, 2\rho_{\boldsymbol{\theta}_\star}(\mathbf{x}_\ell)-1) - a(\mathbf{x}_\ell)(2\rho_{\boldsymbol{\theta}_\star}(\mathbf{x}_\ell) - 1)
\end{equation*}
where we denote by $a(\mathbf{x}_{\ell}) \in \{0,1\}$ the indicator of whether the learner has decided to accept (1) or reject (0) the data point $\mathbf{x}_{\ell}$. 

Minimizing regret is a standard objective in the online learning and bandits literature (see~\cite{lattimore2020bandit}).

Using the set of accepted data points, i.e. the \emph{greedy} approach, $\rho_{\boldsymbol{\theta}_t}$ can be obtained by  minimising binary cross entropy loss:
\begin{equation}
\begin{aligned}
\label{eq:regretloss}
    \mathcal{L}(\boldsymbol{\theta} | \mathcal{A}_t) &= \sum_{\mathbf{x}\in \mathcal{A}_t}    -y(\mathbf{x}) \log\left(\rho_{\boldsymbol{\theta}}(\boldsymbol{x}) \right)\\&-(1-y(\mathbf{x}))\log\left(1-\rho_{\boldsymbol{\theta}}(\mathbf{x})\right) 
\end{aligned}
\end{equation}
where $\mathcal{A}_t = \{ \mathbf{x}_\ell| a_\ell = 1 \text{ and } \ell \leq t-1\}$ denotes the dataset of accepted points up to the time $t-1$. In what follows we call such  $\rho_{\boldsymbol{\theta}_t}$ the \emph{biased model}.

\begin{figure*}
  \centering
  {\includegraphics[width=1\linewidth]{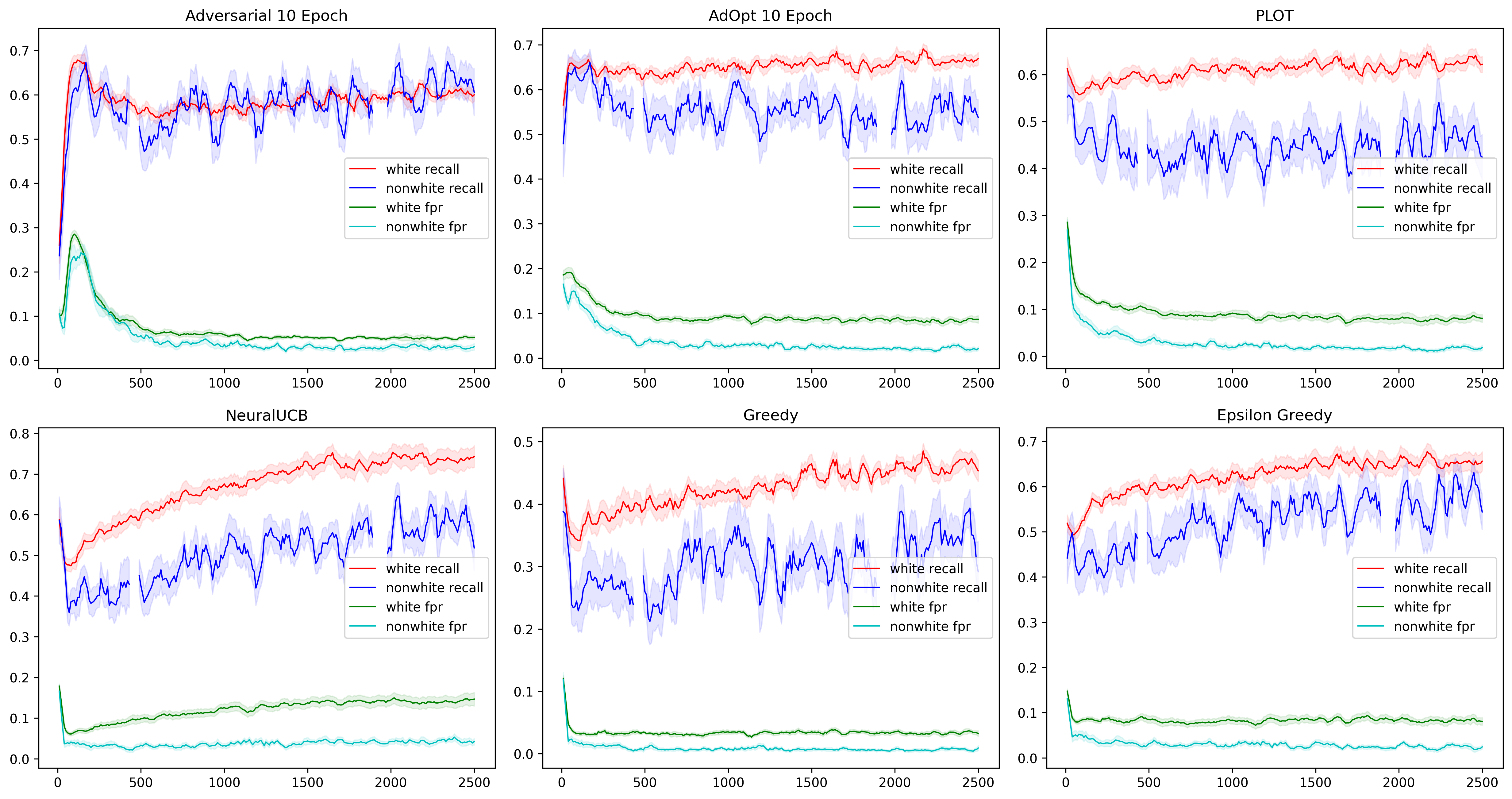}}
  \caption{The recall and false positives ratio (fpr) values per batch for ``White'' and ``Other'' subgroups of the \emph{Census Income(``Adult'')} dataset. The adversarially de-biased classifier is the only one among evaluated methods that shows similar values of recall (tpr) and fpr for these groups, thus coming close to satisfying the equality of odds definition of fairness.}
  \label{fig:fairness}
\end{figure*}

\subsection{Adversarial domain adaptation}
\label{sec:domadapt}
Domain adaptation studies strategies for learning in the presence of distributional shift between the labelled data used for training and the unlabelled data domain containing test samples. Adversarial domain adaptation, first proposed in~\cite{Ganin2015}, is based on the theory of distance between domains introduced in~\cite{BenDavid2006, BenDavid2010} and the GAN approach of~\cite{goodfellow2014generative}.  The main idea of this method is constructing a representation of the data that is both discriminative for the given classification task and domain invariant. 
Further work in adversarial domain adaptation has generalised the method to different adversarial loss functions and generative tasks, e.g.~\cite{Liu2016Coupled, Tzeng2017,zhang2018mitigating}, and successfully applied it to a wide array of challenges, e.g.~\cite{ Edwards2015Censoring,Wang2019}. Our approach in the present work is similar to the DANN algorithm of~\cite{Ganin2015}, that we found to give stable results for the duration of the training.
\begin{table*}[t]
\
   \begin{tabular}{lllllllll}
    \toprule
    {} &
    {} &
    \multicolumn{2}{c}{Recall} &
    \multicolumn{2}{c}{Precision} &
    \multicolumn{3}{c}{Predicted Positive} 
    \\
    \cmidrule(r){3-9}
     &True Pos.& biased     & de-biased &
    biased     &de-biased &
    biased     &de-biased  &de-biased $\mathcal{S}$\\
    \midrule
    Mean &0.24 & 0.59& 0.83 &0.78   &0.31 &0.18  &0.66  & 0.69\\
    STD &0.07 &0.2  & 0.16 &0.19  &0.12 &0.07  &0.11  & 0.11  \\
    \bottomrule
  \end{tabular}
    \centering
  \caption{Mean and STD values for performance metrics of biased (greedy) and adversarially de-biased classifiers on the \emph{Census Income (``Adult'')} dataset. We find that adversarial de-biasing trades off precision and recall. All columns except the last contain metrics for the incoming batch, or ``Target'' $\mathcal{T}$. The last three columns contain the proportion of predicted positives for the incoming batch and the ``Source'' $\mathcal{S}$ set of accepted applicants to empirically validate that equation~(\ref{eq:equidist}) holds.
  }
  \label{tbl:adv-metrics}

\end{table*}

Given a labelled \emph{source} dataset $\mathcal{S}\subset \R^d$ and a \emph{target} dataset $\mathcal{T} \subset \R^d$ whose labels we need to predict, the adversarial domain adaptation approach is to simultaneously train a generator $G_{\theta}:\R^d\rightarrow\R^{d'}$ that encodes the data, a classifier $C_{\psi}:\R^{d'}\rightarrow\{0,1\}$, and a discriminator $D_{\phi}:\R^{d'}\rightarrow\{0,1\}$ that discriminates between the encodings of samples from $\mathcal{S}$ and $\mathcal{T}$, so that $C_{\psi}(G_{\theta}(\cdot))$ has a high classification accuracy on $\mathcal{S}$, and $G_{\theta}$ is trained adversarially with $D_{\phi}$ to minimise the possibility of distinguishing between samples from $G_{\theta}(\mathcal{S})$ and $G_{\theta}(\mathcal{T})$. Following~\cite{zhang2018mitigating} we will refer to the representation of the data by $G_{\theta}$ as ``de-biased'' and to $C_{\psi}$ as the ``de-biased classifier'' - ``de-biased'' in our context refers to a representation that doesn't distinguish between test and train domains.

The training of generator, classifier, and discriminator can be summarised by the training objective $\min_{\theta}\min_{\psi}\max_{\phi}L_{\theta, \psi,\phi}$,
\begin{equation}
\begin{aligned}
    \label{eq:advTraining}
     L_{\theta, \psi,\phi}&:=\mathop{\mathbb{E}}_{\mathbf{x} \in\mathcal{T}} \log(D_{\phi}(G_{\theta}(\mathbf{x})) \\&+ \mathop{\mathbb{E}}_{\mathbf{x}\in\mathcal{S}} \log(1-D_{\phi}(G_{\theta}(\mathbf{x}))) \\&+ L^{C}_{x\in \mathcal{S}}(C_{\psi}(G_{\theta}(\mathbf{x})),y(\mathbf{x})),
\end{aligned}
\end{equation}
where $L^C$ is the binary cross entropy loss and $y:\mathcal{S}\rightarrow \{0,1\}$ provides the labels.

\section{Adversarial domain adaptation approach to the Bank Loan Problem}

\subsection{Adversarial de-biasing in the BLP context} 
\label{par:intuition}
Let us recap the task at hand: in the BLP setting at time $t$ we possess a labelled set of previously accepted applicants $\mathcal{A}_t$ and we want to utilize this information to label a new applicant. The obstacle to overcome is that the distribution of the data points in the general applicant pool $\mathcal{A}$, is different from the distribution of data points in $\mathcal{A}_t$, that was influenced by the previous choices of the classifier. This will compromise the performance of a classifier trained on $\mathcal{A}_t$. We address this issue by using domain adaptation to find the de-biased representation $G_{\theta}$ of the data in $\mathcal{A}_t$ and train the de-biased classifier $C_{\phi}$ to predict a label for the candidate $\mathbf{x}$ sampled from $\mathcal{A}$.

We expect $C_{\phi}$ to give \emph{optimistic} predictions for the labels of $\mathbf{x} \in \mathcal{A}$, in the following sense: as shown in~\cite{Edwards2015Censoring}, adversarial domain adaptation produces a classifier and generator that converge to approximating the \emph{demographic parity} property
\begin{align}
 \label{eq:equidist}
    P_{(C(G(\mathbf{x}))=1\mid \mathbf{x}\in \mathcal{S})}=P_{(C(G(\mathbf{x}))=1\mid \mathbf{x}\in \mathcal{T})}.
\end{align}

Equation~(\ref{eq:equidist}) states that the percentage of positive predictions on the transformed labelled set $G_{\theta}(\mathcal{S})$ is the same as on $G_{\theta}(\mathcal{T})$. In the BLP setting, as the training progresses and the classifier becomes better at accepting the right candidates, the distribution of true positive labels in $\mathcal{A}_t$ diverges from that of the general population: for $t\gg 0$ 
\begin{align}
\label{eq:optgrow}
P_{(y(\mathbf{x})=1|\mathbf{x}\in\mathcal{A}_t)}> P_{(y(\mathbf{x})=1|\mathbf{x}\in\mathcal{A})}.
\end{align}
Note that the above is simply a consequence of the classifier being successful in its job and is independent of the accumulation of bias due to erroneous rejections. Since in the adversarial adaptation approach the classifier $C_{\psi}$ is trained to preserve the accuracy on the labelled set $\mathcal{A}_t$, $P(C_{\psi}(G_{\theta}(\mathbf{x}))=1)$ is going to be close to $P(y_{true}(\mathbf{x})=1|\mathbf{x}\in\mathcal{A}_t)$ \emph{both} when $\mathbf{x}\in\mathcal{A}_t$ \emph{and} $\mathbf{x}\in \mathcal{A}$. In the BLP setting, the adversarial training regime creates de-biased classifier that optimistically overpredicts positive labels for the incoming queries. 

To verify empirically that our adversarial training regime in the context of BLP produces de-biased classifier with properties as above, we create a performance metrics log across the experiments comparing the performance of the de-biased classifier and the classifier trained on the original biased data. The mean and standard deviation for the de-biased classifier that is trained from scratch at each step are in Table~\ref{tbl:adv-metrics}. The last two columns empirically evaluate Equation~(\ref{eq:equidist}).

For practical realisation to increase computational efficiency we consider the scenario, where at each time-step a learner receives \emph{a batch} of data points $\mathcal{B}_t=\{\mathbf{x}_{t,1}, \ldots, \mathbf{x}_{t, B}\}$. We obtain the pair consisting of de-biased classifier and the generator of de-biased data representation $(C,G)$ for the pair of source and target datasets $(\mathcal{A}_t,\mathcal{B}_t)$ by minimising the loss function~(\ref{eq:advTraining}) and use $C$ to obtain predictions for the labels of the data points in $\mathcal{B}_t$. We provide the details of the adversarial training algorithm in the Appendix.

As can be seen from Table~\ref{tbl:adv-metrics}, increased recall of the adversarially de-biased classifier comes at the expense of lower precision. This trade off can be regulated by the length of adversarial training. We empirically found that for minimizing regret best performance is achieved when training for small number of epochs at each step without resetting the weights of the adversarial triad. This approach also reduces the training duration. 

\section{Adversarial optimism}
\label{sec:adopt}

The adversarially de-biased classifier fails to reliably achieve the optimal results on all datasets (Figure~\ref{fig:regrets}), due to high false-positive rate and the inherent instability of the adversarial training, which causes the wide standard deviation on the regret plots. To mitigate for these drawbacks, we propose a new \emph{Adversarial Optimism} (AdOpt) method, that combines the adversarially de-biased classifier with the \emph{pseudo-label optimism} PLOT approach from~\cite{PLOT}.

The PLOT approach is a simple yet efficient way to introduce optimism into a classification routine, which can be used with deep neural networks. The approach can be summarized as follows: to each new query $\mathbf{x}_0$ \emph{rejected} by the ``biased'' classifier  trained on \textit{all the accepted data} so far it optimistically assigns a positive \emph{pseudo-label}, retrains the model on the dataset that includes this data point with positive label and uses the resulting \emph{pseudo-label} model to confirm or ignore the optimistic prediction.

Our proposed method, AdOpt uses two classifiers: the ``biased'' classifier and the adversarially de-biased one. Presented with a new data point AdOpt proceeds as follows:
\begin{itemize}
\item If a data point is \textit{accepted} by the biased classifier, it is accepted and added to the dataset with the true label.
\item If a data point is instead \emph{rejected} by the biased classifier we use the de-biased classifier to decide whether to add it to the pseudo-label dataset to decide whether we should \textit{reconsider} it for acceptance.
\item We then apply the pseudo-label mechanism from PLOT, i.e. retraining on \textit{optimistic labels}, on these candidates to decide final acceptance. 
\end{itemize}
The last step of the algorithm evaluates the points that are recommended for acceptance by the de-biased classifier against the models prior knowledge by adding it to the original dataset with positive pseudo-labels and retraining an optimistic classifier to make a final decision. Incorporating prior model knowledge via PLOT mitigates against the high false acceptance rate, as well as the instability of adversarial training across batches and thus helps to further reduce regret incurred by our algorithm.
\begin{figure*}[t]
  \centering
{\includegraphics[width=1\linewidth]{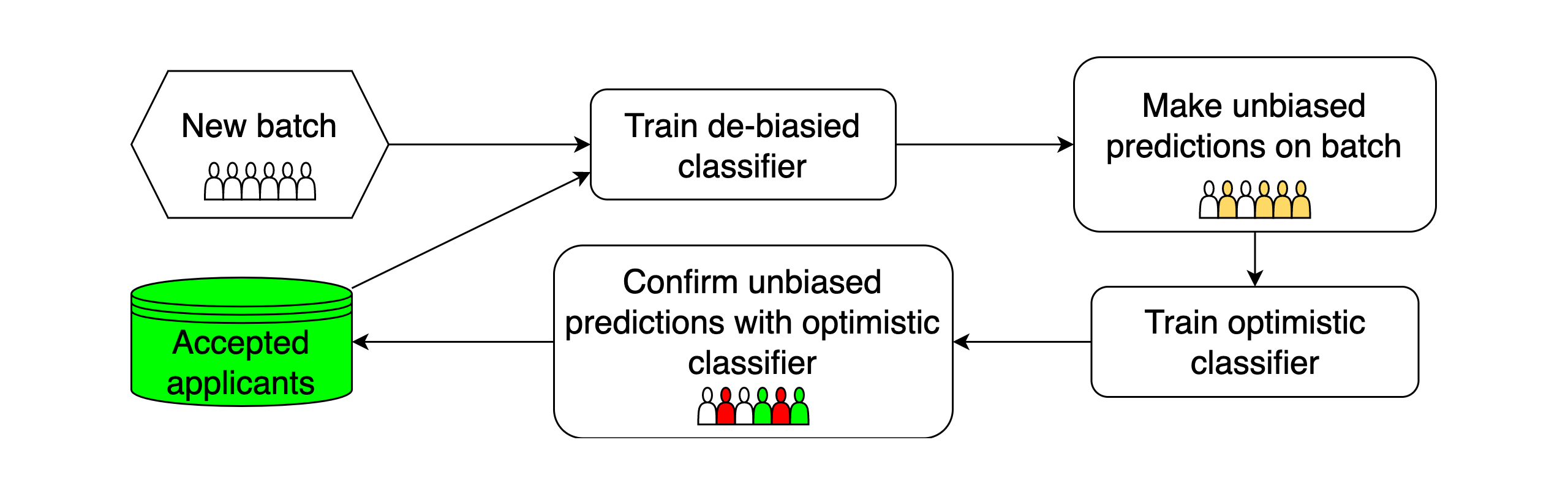}}
  \caption{The AdOpt algorithm}
  \label{fig:algo_diagram}
\end{figure*}

Using the de-biased classifier for assigning pseudo-labels enables AdOpt to explore faster. The de-biased classifier possesses increased recall. While this comes at expense of lower precision, its acceptance suggestions are superior to randomly picking pseudo-label candidates as is done in~\cite{PLOT}. 

As before, in practice at every step we consider a batch of data points $\mathcal{B}_t$. During the very first time-step ($t=1$), the algorithm accepts all the points in $\mathcal{B}_1$. In subsequent time-steps we train a de-biased classifier adversarially for every timestep on the pair $(\mathcal{A}_t,\mathcal{B}_t)$. We identify a subset of the current batch $\mathcal{B}_t^0\subset\mathcal{B}_t$ composed of those points that are both currently being predicted as rejects by the biased model and have been selected as potential positives by the de-biased classifier. They comprise the pseudo-label batch of potential candidates for acceptance. We train the pseudo-label model on the combined dataset and accept those data points, whose labels were confirmed by the pseudo-label model predictions.
The pseudo-label loss is given by adding a term to the loss function:
\begin{equation}
\begin{aligned}
\label{eq:AdOptLoss}
    \mathcal{L}^\mathcal{C}(\boldsymbol{\theta} | \mathcal{A}_t, \mathcal{B}_t) &= \mathcal{L}(\boldsymbol{\theta} | \mathcal{A}_t \cup \{ (\mathbf{x}, 1) \text{ for } \mathbf{x} \in \mathcal{B}_t \} ) \\
    &=\sum_{\mathbf{x}\in \mathcal{A}_t} -y\log\left(\rho_{\boldsymbol{\theta}}(\mathbf{x})\right) \\&- (1-y) \log(1-\rho_{\boldsymbol{\theta}}(\mathbf{x})) \\&- 
    \underbrace{ \sum_{\boldsymbol{x} \in \mathcal{B}_t^0} \log(\rho_{\boldsymbol{\theta}}(\mathbf{x}))}_{\text{pseudo-label loss}},
\end{aligned}
\end{equation}
\begin{algorithm}[ht]
\caption{AdOpt}
\label{alg:AdvOpt}
Initialize accepted dataset $\mathcal{A}_1 = \emptyset$ \\
For {$t=1, \cdots, T$}\\{
    1. Observe batch $\mathcal{B}_t = \{ \mathbf{x}_t^{(j)}\}_{j=1}^B$.\\
    2. Run adversarial domain adaptation algorithm with starting weights initialized from training $\mathcal{B}_{t-1}$ on $(\mathcal{A}_t,\mathcal{B}_t)$ to obtain generator and classifier pair $(G,C)$.\\
    3. Train the biased model  $\rho_{\boldsymbol{\theta}_t}$ by minimizing the loss $\mathcal{L}_t(\boldsymbol{\theta} | \mathcal{A}_t)$.\\
    4. Compute the pseudo-label filtered batch $\mathcal{B}_t^0 = \{ (\mathbf{x}_t^{(j)}, 1) \mid \rho_{\boldsymbol{\theta}_t}(\mathbf{x}_t^{(j)} ) < \frac{1}{2} \text{ and } C(G(x_t^j))=1 \}$. \\
    5. Train the pseudo-label model by minimizing the optimistic pseudo-label loss, $\mathcal{L}_t^{\mathcal{C}}(\boldsymbol{\theta} | \mathcal{A}_{t} , \widetilde{\mathcal{B}}_t)$\\
    6. For $\mathbf{x}^{(j)} \in \mathcal{B}_t$ compute acceptance decision via 
    $a^{(j)}_t = \begin{cases}1 &\text{if } \rho_{\boldsymbol{\theta}^C_t}(\boldsymbol{x}_t) \geq \frac{1}{2} \text { or } \rho_{\boldsymbol{\theta}_t}(\boldsymbol{x}_t) \geq \frac{1}{2}\\
    0 &\text{o.w.}\end{cases}$ \\
    7. Update $\mathcal{A}_{t+1} \leftarrow \mathcal{A}_t \cup \{ (\mathbf{x}^{(j)}_t) \}_{ j \in \{1, \cdots, B\} \text{ s.t. } a_t^{(j)} = 1} $.}
\end{algorithm}

A notable feature of PLOT is that as the size of the dataset of accepted points grows, the predictions of the pseudo-label model differ less and less from the predictions of the biased model trained on the set of the accepted points. Ideologically once the dataset is sufficiently large and accurate information can be inferred about the true labels, the inclusion of $\mathcal{B}_t^0$ into the pseudo-label loss has vanishing effect on the model's predictions. The latter has the beneficial effect of making false positive rate decrease with $t$. This balances the opposite trend in the predictions of the de-biased adversarial classifier, that tends to predicting more positives as the training progresses as a result of the Equations~(\ref{eq:optgrow}) and~(\ref{eq:equidist}).

\subsection{Fairness aspects}
The datasets considered in the BLP context can explicitly or implicitly reveal sensitive information, such as e.g. gender or race. It is therefore natural to ask to what extent the methods in the BLP domain satisfy fairness criteria. While a comprehensive evaluation with respect to the multitude of definitions of fairness~\cite{verma2018fairness,hertweck2022gradual} would be well beyond of the scope of our paper, both intuition and prior works~\cite{Edwards2015Censoring, zhang2018mitigating} suggest that de-biased classifier should perform well for at least some of these metrics. To elaborate, consider for example the positively labelled (i.e. ``persons with income greater than USD 50K'') subset of~\emph{Census Income(``Adult'')} dataset. We find that males and Whites are overrepresented compared to females and representatives of all other races respectively. It follows that in BLP setting the classifier will encounter significantly less positively labelled females and non-Whites than males and Whites, and the self-reinforcing false rejections phenomenon would make it prone to acquiring negative bias towards these groups. The above analysis shows that fairness investigation is warranted in this setting. Adversarial de-biasing should be able to partially mitigate this effect, because it uses a representation of the data which doesn't distinguish between the accepted applicants and the applicants in the current batch. We verify this effect empirically in Figure~\ref{fig:fairness}: our experiments on the ``Adult'' dataset show that adversarially de-biased classifier comes close to satisfying the \emph{equality of odds} definition of fairness: namely the true positive rate (TPR) or ``recall'' and the false positive rate (FPR) are close for the ``Whites'' and ``Others'' subgroups of the dataset. In the meanwhile the leading approaches to BLP show markedly different results for these subgroups. The intuition behind the equality of odds definition in the Bank Loan Problem language can be formulated as a requirement that
\begin{itemize}
\item Credible customers from both groups should be awarded similar opportunities.
\item No unfair opportunity advantage should be given to those customers that are going to default for either group.
\end{itemize}
The results for the female/male subgroups are similar and are presented in Figure~\ref{fig:gender}. We note that incorporation of adversarially de-biased classifier in AdOpt improves its performance on this metric compared to PLOT.

We defer the more through investigation of fairness aspects to future work.
\section{Experiments}
\label{sec:results}

\paragraph{Datasets and methods}
To assess the performance of adversarially de-biased classifier and AdOpt, we consider several benchmark datasets: the UCI datasets \emph{Census 
Income(``Adult'')} and \emph{Default of Credit Card Clients}~\cite{lichman2013uci}, as well as the anonymized dataset of home equity line of credit (HELOC)~\cite{FICO} and MNIST converted to the binary reward format of the BLP by defining the positive class to be images of the digit 5. ``Adult'' contains demographic information useful for fairness evaluation, and was used in in~\cite{PLOT} alongside MNIST, while FICO and ``Credit'' are publicly available real life datasets used for financial credibility evaluation. We train adversarially de-biased classifier for 10 epochs at each step.

We compare against four baseline methods: PLOT, NeuralUCB, Greedy (no exploration), and decayed $\epsilon$-greedy method. For $\epsilon$-greedy, we use a decayed schedule of \cite{kveton2019garbage}, dropping to 0.1\% exploration by T=2500. We combine PLOT with $\epsilon$-greedy selection of pseudo-label candidates as in \cite{PLOT}. 

Since sampling of batches presented during training as well as stochastic nature of optimisation has a significant effect on the final regret, to obtain the results in this paper we average the performance of each method across 5 different batch sampling methods and 5 separate runs of 2500 timesteps each, so altogether 25 experiments for each method. We note that this is different from the approach in ~\cite{PLOT} where only one specific way of sampling examples from the datasets was examined. This explains the difference in performance for PLOT on MNIST reported in this paper and in ~\cite{PLOT}. 
We also note that, as opposed to ~\cite{PLOT}, we report absolute values of regret at every step, rather than regret with respect to a baseline model, which allows us to achieve more transparent results. The high number of experiments allows us to conduct t-test and compute $0.95$ confidence intervals to evaluate statistical significance of reported results.

The number of training epochs for the adversarially de-biased classifier is a parameter that can be tuned to optimise the results for each dataset, however we chose to use fixed number of epochs for our comparisons in this paper, as it was already sufficient to demonstrate statistically significant advantage over the SOTA. The number of epochs also influences the runtime of the AdOpt algorithm.  However we prioritise lower regret over faster runtime as for our purposes compute is cheap, while mistakes are costly. Further particulars on experiment details and hyperparameters can be found in appendix.

\paragraph{Main results}
\begin{figure}
  \centering
  {\includegraphics[width=\linewidth]{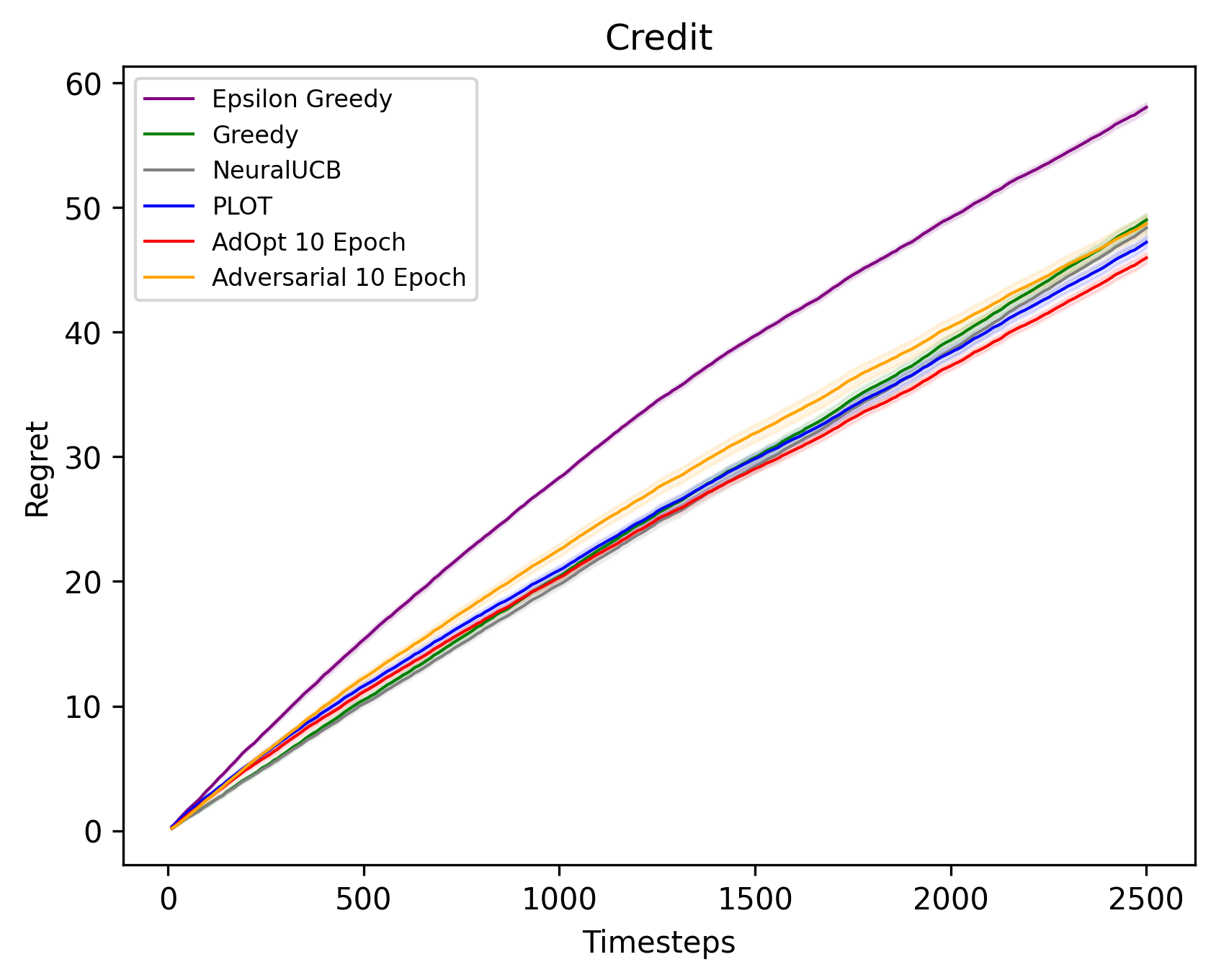}}
  \caption{Cumulative regret of adversarially de-biased classifier (``Adversarial'') and AdOpt  vs. PLOT, NeuralUCB, greedy and $\epsilon$-greedy methods on the UCI dataset Default of Credit Card Clients (``Credit'').
  }
  \label{fig:regrets_credit}
\end{figure}
We report the following main results: 
We experimentally evaluate the performance of our proposed algorithm, AdOpt, that combines the adversarially de-biased classifier with the pseudo-labelling approach of~\cite{PLOT}. We find that AdOpt significantly outperforms previously proposed methods. In Figures~\ref{fig:regrets} and~\ref{fig:regrets_credit} we report cumulative regrets for each method.  

\begin{figure}
  \centering
  {\includegraphics[width=1\linewidth] {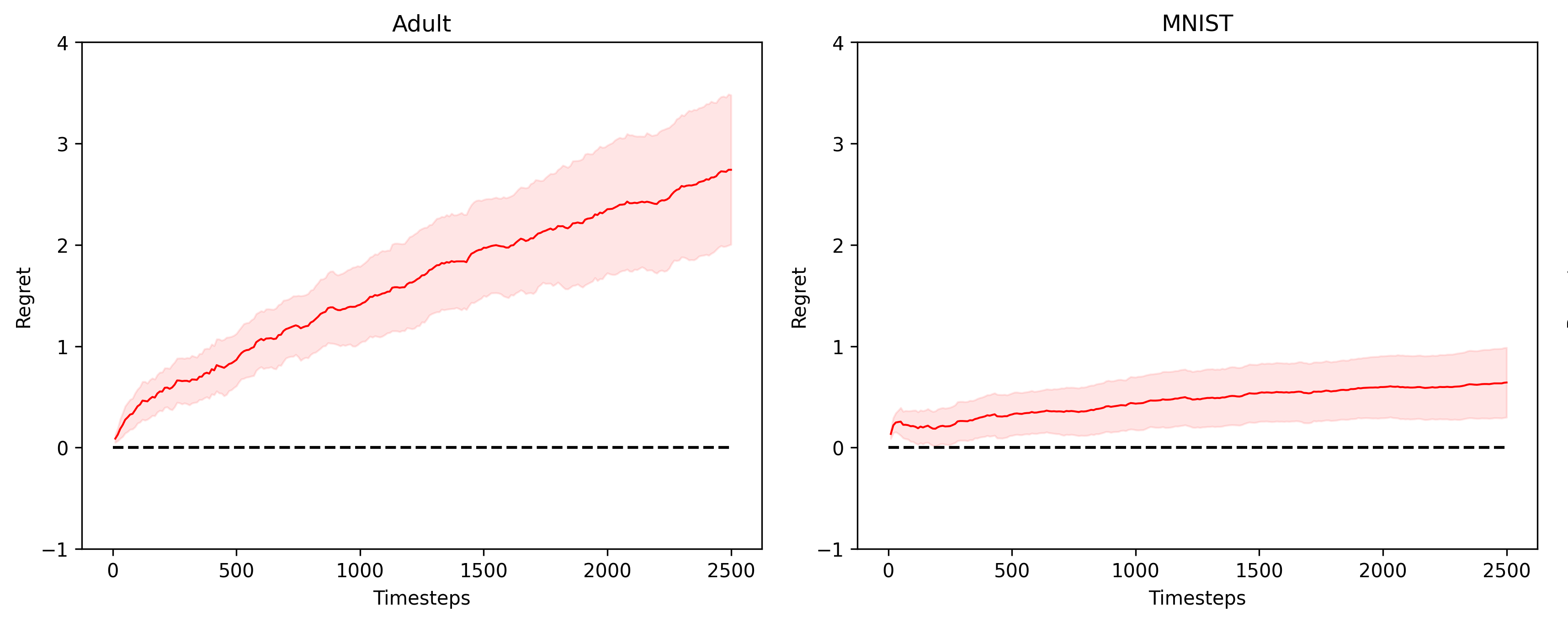}}
  \caption{The difference of cumulative regrets for PLOT and AdOpt with $0.95$ confidence interval shaded on Census Income and MNIST datasets.}
  \label{fig:PLOTAdOpt}
\end{figure}

To evaluate the significance of improvement of AdOpt over PLOT we compute paired t-values for each dataset in Table~\ref{tbl:tvalues}. 
\begin{table}
  \centering
  \vskip 0.15in
  \begin{tabular}{lllllll}
    \toprule
    & Adult & FICO     & MNIST      &Credit  \\
    \midrule
    AdOpt &6.8798 &6.7142 &4.0966  &4.1172 \\
    \bottomrule
  \end{tabular}
  \caption{Paired t-values of cumulative regret of AdOpt vs previous SOTA method PLOT at 2500 steps, where t-values of $>2$ indicate $<0.05$ chance of null-hypothesis. 
  }
\label{tbl:tvalues}
\end{table}
In Figure~\ref{fig:PLOTAdOpt} we display the corresponding $0.95$ CI for the difference between cumulative regrets of PLOT and AdOpt on ``Adult'' and MNIST datasets.

In Table~\ref{tbl:adv-metrics}, we compare the performance metrics for the adversarially de-biased classifier and the greedy classifier trained on the original biased data on the ``Adult'' dataset. The de-biased classifier achieves higher recall while maintaining predictive ability evidenced by its precision value. 

We conduct experiments to evaluate fairness for the methods in the BLP domain. The results for the Census Income dataset are shown in Figures~\ref{fig:fairness} and~\ref{fig:gender}. Our experiments indicate that the adversarially de-biased classifier exhibits significantly less bias in its predictions for different subgroups within those categories, in particular displaying similar TPR and FPR values, and its incorporation within our proposed method, AdOpt, improves performance with respect to these metrics.

\section{Conclusion}
We apply adversarial domain adaptation to online learning to address the distributional shift accumulating due to past acceptance or rejection decisions. Our empirical evaluation indicates that the adversarially de-biased classifier has the potential to greatly reduce the risk of \textit{self-reinforcing rejection-loops}, as well as increase fairness towards minority subgroups. We present AdOpt, a novel algorithm that incorporates adversarial domain adaptation and pseudo-label optimism and experimentally establishes a new SOTA across a variety of instances of the BLP with minimal hyperparameter tuning. Further directions include extensive exploration of other fairness aspects for online application of domain adaptation based methods, as well as generalisation of AdOpt to higher dimensional actions supporting applications to a variety of datasets and problem settings.
\appendix
\section{Experiment details and hyperparameters}
\label{app:hyperparameters}
Our experiments use the optimal hyperparameters for each method reported in previous papers. NeuralUCB uses an alpha of $0.4$, and a discount factor of $0.9$. PLOT uses a two-layer, 40-node, multi-layer perceptron. AdOpt is trained using an encoded dimension of 100, as well as a hidden layer size of 100. Each experiment runs on a single GPU, and our results were collected by trivially distributing 25 experiments in parallel on 25 different GPU cores. 

For the adversarial training, the generator, classifier, and discriminator are multi-layer perceptrons, where the dimension of the input, hidden layer(s), and output are $[\text{Data Dimension}, 100, 100, 100]$, $[100, 100, 1]$, and $[100, 100, 100, 1]$ respectively. We used Adam optimizer, with $(\beta_1,\beta_2)=(0,0.99)$ and a learning rate of $0.0005$ for the discriminator and $0.0001$ for the generator and classifier \cite{Adam2014}. 

We use the batch size of $32$. As the historical dataset of accepted points grows, there is a large disparity in size between the sample (new batch) and target (historical) datasets for adversarial domain adaptation. To handle this data size discrepancy, we restrict the target dataset for adversarial training to a randomly sampled batch of size $32*100$ from the historical data, once the data reaches this size. As we use a batch size of $32$, this occurs after 100 time steps.

\section{Acknowledgments}
The authors would like to acknowledge the use of the University of Oxford Advanced Research Computing (ARC) facility in carrying out this work. http://dx.doi.org/10.5281/zenodo.22558

Elena Gal and Ben Walker were funded by the Hong Kong Innovation and Technology Commission (InnoHK Project CIMDA).

Aldo Pacchiano was supported by funding from the Eric and Wendy Schmidt Center at the Broad Institute of MIT and Harvard.

Terry Lyons was funded in part by the EPSRC [grant number EP/S026347/1], in part by The Alan Turing Institute under the EPSRC grant EP/N510129/1, the Data Centric Engineering Programme (under the Lloyd’s Register Foundation grant G0095), the Defence and Security Programme (funded by the UK Government) and the Office for National Statistics and The Alan Turing Institute (strategic partnership) and in part by the Hong Kong Innovation and Technology Commission (InnoHK Project CIMDA).

\bibliography{main}

\begin{thebibliography}{38}
\providecommand{\natexlab}[1]{#1}

\bibitem[{Abbasi-Yadkori, P{\'a}l, and
  Szepesv{\'a}ri(2011)}]{abbasi2011improved}
Abbasi-Yadkori, Y.; P{\'a}l, D.; and Szepesv{\'a}ri, C. 2011.
\newblock Improved algorithms for linear stochastic bandits.
\newblock \emph{Advances in neural information processing systems}, 24.

\bibitem[{Auer et~al.(2002)Auer, Cesa-Bianchi, Freund, and
  Schapire}]{auer2002nonstochastic}
Auer, P.; Cesa-Bianchi, N.; Freund, Y.; and Schapire, R.~E. 2002.
\newblock The nonstochastic multiarmed bandit problem.
\newblock \emph{SIAM journal on computing}, 32(1): 48--77.

\bibitem[{Ben-David et~al.(2010)Ben-David, Blitzer, Crammer, Kulesza, Pereira,
  and Vaughan}]{BenDavid2010}
Ben-David, S.; Blitzer, J.; Crammer, K.; Kulesza, A.; Pereira, F.; and Vaughan,
  J. 2010.
\newblock A theory of learning from different domains.
\newblock \emph{Machine Learning}, 79: 151--175.

\bibitem[{Ben-David et~al.(2006)Ben-David, Blitzer, Crammer, and
  Pereira}]{BenDavid2006}
Ben-David, S.; Blitzer, J.; Crammer, K.; and Pereira, F. 2006.
\newblock Analysis of Representations for Domain Adaptation.
\newblock In Sch\"{o}lkopf, B.; Platt, J.; and Hoffman, T., eds.,
  \emph{Advances in Neural Information Processing Systems}, volume~19. MIT
  Press.

\bibitem[{Coston et~al.(2019)Coston, Ramamurthy, Wei, Varshney, Speakman,
  Mustahsan, and Chakraborty}]{coston2019fair}
Coston, A.; Ramamurthy, K.~N.; Wei, D.; Varshney, K.~R.; Speakman, S.;
  Mustahsan, Z.; and Chakraborty, S. 2019.
\newblock Fair transfer learning with missing protected attributes.
\newblock In \emph{Proceedings of the 2019 AAAI/ACM Conference on AI, Ethics,
  and Society}, 91--98.

\bibitem[{Coston, Rambachan, and Chouldechova(2021)}]{coston2021characterizing}
Coston, A.; Rambachan, A.; and Chouldechova, A. 2021.
\newblock Characterizing fairness over the set of good models under selective
  labels.
\newblock In \emph{International Conference on Machine Learning}, 2144--2155.
  PMLR.

\bibitem[{Edwards and Storkey(2016)}]{Edwards2015Censoring}
Edwards, H.; and Storkey, A.~J. 2016.
\newblock Censoring Representations with an Adversary.
\newblock In \emph{ICLR 2016}.

\bibitem[{FICO. Explainable machine learning challenge()}]{FICO}
FICO. Explainable machine learning challenge. 2018.
\newblock \url{https://community.fico.com/
  s/explainable-machine-learning-challenge.}
\newblock Accessed: 2022-01-01.

\bibitem[{Filippi et~al.(2010)Filippi, Cappe, Garivier, and
  Szepesv\'{a}ri}]{glm_bandit}
Filippi, S.; Cappe, O.; Garivier, A.; and Szepesv\'{a}ri, C. 2010.
\newblock Parametric Bandits: The Generalized Linear Case.
\newblock In Lafferty, J.~D.; Williams, C. K.~I.; Shawe-Taylor, J.; Zemel,
  R.~S.; and Culotta, A., eds., \emph{Advances in Neural Information Processing
  Systems 23}, 586--594.

\bibitem[{Foster and Rakhlin(2020)}]{foster2020beyond}
Foster, D.; and Rakhlin, A. 2020.
\newblock Beyond UCB: Optimal and efficient contextual bandits with regression
  oracles.
\newblock In \emph{International Conference on Machine Learning}, 3199--3210.
  PMLR.

\bibitem[{Ganin et~al.(2016)Ganin, Ustinova, Ajakan, Germain, Larochelle,
  Laviolette, Marchand, and Lempitsky}]{Ganin2015}
Ganin, Y.; Ustinova, E.; Ajakan, H.; Germain, P.; Larochelle, H.; Laviolette,
  F.; Marchand, M.; and Lempitsky, V. 2016.
\newblock Domain-Adversarial Training of Neural Networks.
\newblock \emph{J. Mach. Learn. Res.}, 17(1): 2096–2030.

\bibitem[{Goodfellow et~al.(2014)Goodfellow, Pouget-Abadie, Mirza, Xu,
  Warde-Farley, Ozair, Courville, and Bengio}]{goodfellow2014generative}
Goodfellow, I.; Pouget-Abadie, J.; Mirza, M.; Xu, B.; Warde-Farley, D.; Ozair,
  S.; Courville, A.; and Bengio, Y. 2014.
\newblock Generative adversarial networks.
\newblock In \emph{Advances in neural information processing systems},
  2672--2680.

\bibitem[{Hardt et~al.(2016)Hardt, Megiddo, Papadimitriou, and
  Wootters}]{hardt2016strategic}
Hardt, M.; Megiddo, N.; Papadimitriou, C.; and Wootters, M. 2016.
\newblock Strategic classification.
\newblock In \emph{Proceedings of the 2016 ACM conference on innovations in
  theoretical computer science}, 111--122.

\bibitem[{Hashimoto et~al.(2018)Hashimoto, Srivastava, Namkoong, and
  Liang}]{hashimoto2018fairness}
Hashimoto, T.; Srivastava, M.; Namkoong, H.; and Liang, P. 2018.
\newblock Fairness without demographics in repeated loss minimization.
\newblock In \emph{International Conference on Machine Learning}, 1929--1938.
  PMLR.

\bibitem[{Hertweck and R{\"a}z(2022)}]{hertweck2022gradual}
Hertweck, C.; and R{\"a}z, T. 2022.
\newblock Gradual (in) compatibility of fairness criteria.
\newblock In \emph{Proceedings of the AAAI Conference on Artificial
  Intelligence}, volume~36, 11926--11934.

\bibitem[{Joseph et~al.(2016)Joseph, Kearns, Morgenstern, and
  Roth}]{joseph2016fairness}
Joseph, M.; Kearns, M.; Morgenstern, J.~H.; and Roth, A. 2016.
\newblock Fairness in learning: Classic and contextual bandits.
\newblock \emph{Advances in neural information processing systems}, 29.

\bibitem[{Kilbertus et~al.(2020)Kilbertus, Rodriguez, Sch{\"o}lkopf, Muandet,
  and Valera}]{kilbertus2020fair}
Kilbertus, N.; Rodriguez, M.~G.; Sch{\"o}lkopf, B.; Muandet, K.; and Valera, I.
  2020.
\newblock Fair decisions despite imperfect predictions.
\newblock In \emph{International Conference on Artificial Intelligence and
  Statistics}, 277--287. PMLR.

\bibitem[{Kingma and Ba(2014)}]{Adam2014}
Kingma, D.; and Ba, J. 2014.
\newblock Adam: A Method for Stochastic Optimization.
\newblock \emph{International Conference on Learning Representations}.

\bibitem[{Kveton et~al.(2019)Kveton, Szepesvari, Vaswani, Wen, Lattimore, and
  Ghavamzadeh}]{kveton2019garbage}
Kveton, B.; Szepesvari, C.; Vaswani, S.; Wen, Z.; Lattimore, T.; and
  Ghavamzadeh, M. 2019.
\newblock Garbage in, reward out: Bootstrapping exploration in multi-armed
  bandits.
\newblock In \emph{International Conference on Machine Learning}, 3601--3610.
  PMLR.

\bibitem[{Lakkaraju et~al.(2017)Lakkaraju, Kleinberg, Leskovec, Ludwig, and
  Mullainathan}]{lakkaraju2017selective}
Lakkaraju, H.; Kleinberg, J.; Leskovec, J.; Ludwig, J.; and Mullainathan, S.
  2017.
\newblock The selective labels problem: Evaluating algorithmic predictions in
  the presence of unobservables.
\newblock In \emph{Proceedings of the 23rd ACM SIGKDD International Conference
  on Knowledge Discovery and Data Mining}, 275--284.

\bibitem[{Lattimore and Szepesv{\'a}ri(2020)}]{lattimore2020bandit}
Lattimore, T.; and Szepesv{\'a}ri, C. 2020.
\newblock \emph{Bandit algorithms}.
\newblock Cambridge University Press.

\bibitem[{Lichman et~al.(2013)}]{lichman2013uci}
Lichman, M.; et~al. 2013.
\newblock UCI machine learning repository.

\bibitem[{Liu and Tuzel(2016)}]{Liu2016Coupled}
Liu, M.-Y.; and Tuzel, O. 2016.
\newblock Coupled Generative Adversarial Networks.
\newblock In \emph{Proceedings of the 30th International Conference on Neural
  Information Processing Systems}, 469–477. Red Hook, NY, USA: Curran
  Associates Inc.
\newblock ISBN 9781510838819.

\bibitem[{Mancisidor et~al.(2020)Mancisidor, Kampffmeyer, Aas, and
  Jenssen}]{mancisidor2020deep}
Mancisidor, R.~A.; Kampffmeyer, M.; Aas, K.; and Jenssen, R. 2020.
\newblock Deep generative models for reject inference in credit scoring.
\newblock \emph{Knowledge-Based Systems}, 196: 105758.

\bibitem[{Miller, Perdomo, and Zrnic(2021)}]{miller2021outside}
Miller, J.; Perdomo, J.~C.; and Zrnic, T. 2021.
\newblock Outside the Echo Chamber: Optimizing the Performative Risk.
\newblock \emph{arXiv preprint arXiv:2102.08570}.

\bibitem[{Pacchiano et~al.(2021)Pacchiano, Singh, Chou, Berg, and
  Foerster}]{PLOT}
Pacchiano, A.; Singh, S.; Chou, E.; Berg, A.; and Foerster, J. 2021.
\newblock Neural Pseudo-Label Optimism for the Bank Loan Problem.
\newblock \emph{Advances in Neural Information Processing Systems}, 34.

\bibitem[{Perdomo et~al.(2020)Perdomo, Zrnic, Mendler-D{\"u}nner, and
  Hardt}]{perdomo2020performative}
Perdomo, J.; Zrnic, T.; Mendler-D{\"u}nner, C.; and Hardt, M. 2020.
\newblock Performative prediction.
\newblock In \emph{International Conference on Machine Learning}, 7599--7609.
  PMLR.

\bibitem[{Riquelme, Tucker, and Snoek(2018)}]{riquelme2018deep}
Riquelme, C.; Tucker, G.; and Snoek, J. 2018.
\newblock Deep Bayesian Bandits Showdown: An Empirical Comparison of Bayesian
  Deep Networks for Thompson Sampling.
\newblock arXiv:1802.09127.

\bibitem[{Schumann et~al.(2019)Schumann, Lang, Mattei, and
  Dickerson}]{schumann2019group}
Schumann, C.; Lang, Z.; Mattei, N.; and Dickerson, J.~P. 2019.
\newblock Group fairness in bandit arm selection.
\newblock \emph{arXiv preprint arXiv:1912.03802}.

\bibitem[{Sen et~al.(2021)Sen, Rakhlin, Ying, Kidambi, Foster, Hill, and
  Dhillon}]{sen2021top}
Sen, R.; Rakhlin, A.; Ying, L.; Kidambi, R.; Foster, D.; Hill, D.; and Dhillon,
  I. 2021.
\newblock Top-$k$ eXtreme Contextual Bandits with Arm Hierarchy.
\newblock \emph{arXiv preprint arXiv:2102.07800}.

\bibitem[{Singh et~al.(2021)Singh, Singh, Mhasawade, and
  Chunara}]{singh2021fairness}
Singh, H.; Singh, R.; Mhasawade, V.; and Chunara, R. 2021.
\newblock Fairness violations and mitigation under covariate shift.
\newblock In \emph{Proceedings of the 2021 ACM Conference on Fairness,
  Accountability, and Transparency}, 3--13.

\bibitem[{Tzeng et~al.(2017)Tzeng, Hoffman, Saenko, and Darrell}]{Tzeng2017}
Tzeng, E.; Hoffman, J.; Saenko, K.; and Darrell, T. 2017.
\newblock Adversarial Discriminative Domain Adaptation.
\newblock In \emph{2017 IEEE Conference on Computer Vision and Pattern
  Recognition (CVPR)}, 2962--2971. Los Alamitos, CA, USA: IEEE Computer
  Society.

\bibitem[{Verma and Rubin(2018)}]{verma2018fairness}
Verma, S.; and Rubin, J. 2018.
\newblock Fairness definitions explained.
\newblock In \emph{Proceedings of the international workshop on software
  fairness}, 1--7.

\bibitem[{Wang et~al.(2019)Wang, Gan, Liu, Liu, Gao, and Wang}]{Wang2019}
Wang, H.; Gan, Z.; Liu, X.; Liu, J.; Gao, J.; and Wang, H. 2019.
\newblock Adversarial Domain Adaptation for Machine Reading Comprehension.
\newblock In \emph{Proceedings of the 2019 Conference on Empirical Methods in
  Natural Language Processing and the 9th International Joint Conference on
  Natural Language Processing (EMNLP-IJCNLP)}, 2510--2520. Hong Kong, China:
  Association for Computational Linguistics.

\bibitem[{Wei(2021)}]{wei2021decision}
Wei, D. 2021.
\newblock Decision-making under selective labels: Optimal finite-domain
  policies and beyond.
\newblock In \emph{International Conference on Machine Learning}, 11035--11046.
  PMLR.

\bibitem[{Xu et~al.(2020)Xu, Wen, Zhao, and Gu}]{xu2020neural}
Xu, P.; Wen, Z.; Zhao, H.; and Gu, Q. 2020.
\newblock Neural Contextual Bandits with Deep Representation and Shallow
  Exploration.
\newblock \emph{arXiv preprint arXiv:2012.01780}.

\bibitem[{Zhang, Lemoine, and Mitchell(2018)}]{zhang2018mitigating}
Zhang, B.~H.; Lemoine, B.; and Mitchell, M. 2018.
\newblock Mitigating unwanted biases with adversarial learning.
\newblock In \emph{Proceedings of the 2018 AAAI/ACM Conference on AI, Ethics,
  and Society}, 335--340.

\bibitem[{Zhou, Li, and Gu(2020)}]{zhou2020neural}
Zhou, D.; Li, L.; and Gu, Q. 2020.
\newblock Neural Contextual Bandits with UCB-Based Exploration.
\newblock In \emph{Proceedings of the 37th International Conference on Machine
  Learning}, ICML'20. JMLR.org.

\end{thebibliography}

\begin{figure*}
  \centering
  {\includegraphics[width=1\linewidth]{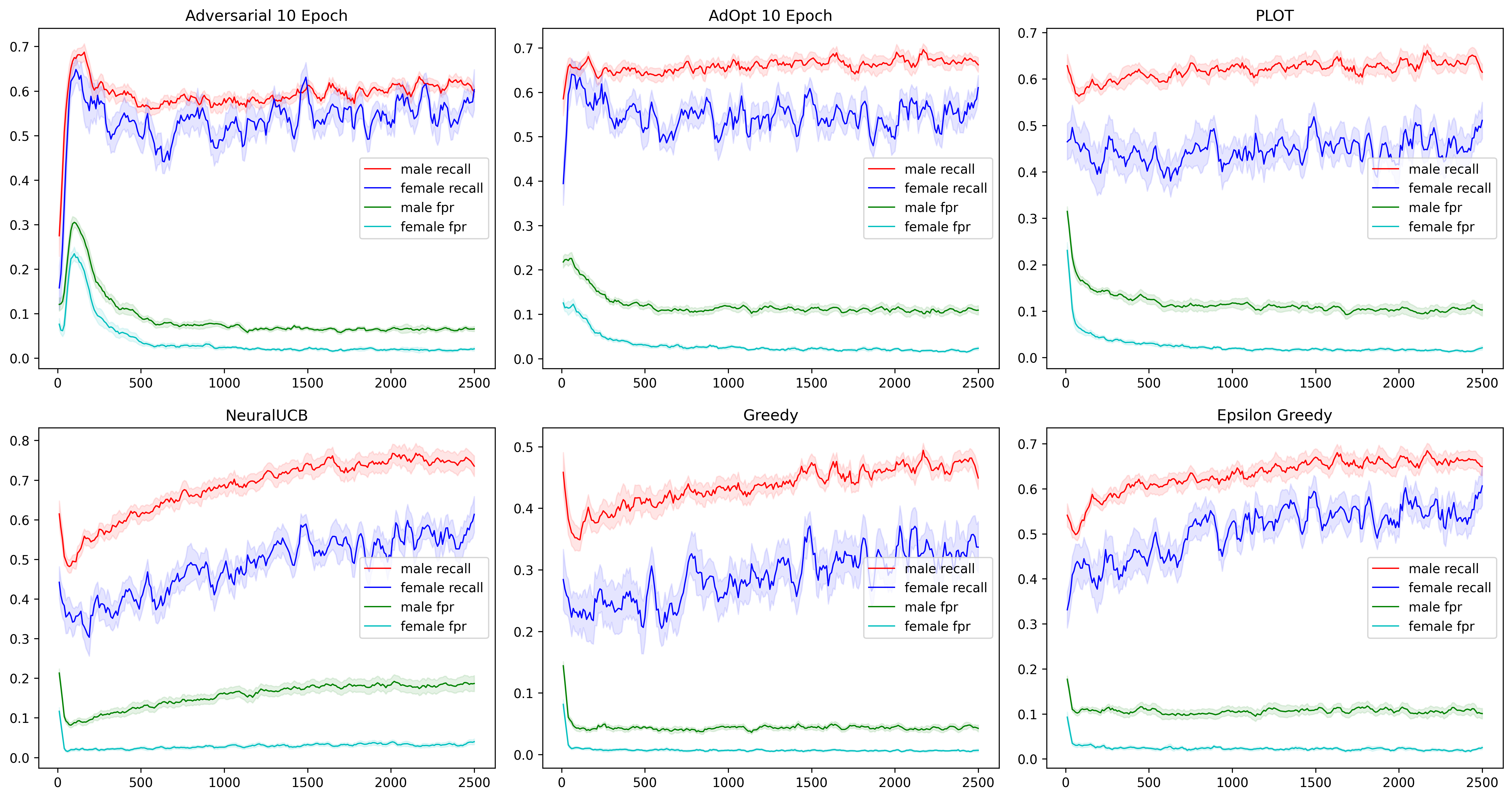}}
  \caption{The recall and false positives ratio (fpr) values per batch for Females and Males subgroups of the ``Census Income'' dataset for the different methods evaluated in the paper. The adversarially de-biased classifier is the only one among evaluated methods that shows similar values of recall and FPR for these groups, thus coming close to satisfying the equality of odds definition of fairness.}
  \label{fig:gender}
\end{figure*}

\end{document}